\documentclass[sigconf,nonacm]{acmart}
\AtBeginDocument{%
  }
\begin{document}
\title{WalkCLIP: Multimodal Learning for Urban Walkability Prediction}
\author{Shilong Xiang}
\email{xiang218@umn.edu}
\affiliation{%
  \institution{University of Minnesota Twin Cities}
  \city{Minneapolis}
  \state{MN}
  \country{USA}}

\author{JangHyeon Lee}
\email{lee04588@umn.edu}
\affiliation{%
  \institution{University of Minnesota Twin Cities}
  \city{Minneapolis}
  \state{MN}
  \country{USA}}

\author{Min Namgung}
\email{namgu007@umn.edu}
\affiliation{%
  \institution{University of Minnesota Twin Cities}
  \city{Minneapolis}
  \state{MN}
  \country{USA}}

\author{Yao-Yi Chiang}
\email{yaoyi@umn.edu}
\affiliation{%
  \institution{University of Minnesota Twin Cities}
  \city{Minneapolis}
  \state{MN}
  \country{USA}}
\renewcommand{\shortauthors}{Xiang et al.}
\begin{abstract}
Urban walkability is a cornerstone of public health, sustainability, and quality of life. Traditional walkability assessments rely on surveys and field audits, which are costly and difficult to scale. Recent studies have used satellite imagery, street view imagery, or population indicators to estimate walkability, but these single-source approaches capture only one dimension of the walking environment. Satellite data describe the built environment from above, but overlook the pedestrian perspective. Street view imagery captures conditions at the ground level, but lacks broader spatial context. Population dynamics reveal patterns of human activity but not the visual form of the environment. We introduce WalkCLIP, a multimodal framework that integrates these complementary viewpoints to predict urban walkability. WalkCLIP learns walkability-aware vision-language representations from GPT-4o generated image captions, refines these representations with a spatial aggregation module that incorporates neighborhood context, and fuses the resulting features with representations from a population dynamics foundation model. Evaluated at 4,660 locations throughout Minneapolis-Saint Paul, WalkCLIP outperforms unimodal and multimodal baselines in both predictive accuracy and spatial alignment. These results show that the integration of visual and behavioral signals yields reliable predictions of the walking environment.
\end{abstract}
\ccsdesc[500]{Computing methodologies~Artificial intelligence}
\keywords{Urban Walkability, Multimodal Representation}
\maketitle
\thispagestyle{plain}
\pagestyle{plain}
\section{Introduction}
Walkability describes how safe, comfortable, and convenient the environment is for residents to reach daily destinations on foot. Empirically, higher walkability is associated with improved health condition~\cite{forsyth2015walkable} and quality of life~\cite{leyden2024walkable}. Conceptually, walkability also functions as a long-term indicator of place-based health risk and resilience, as elements of the built environment can shape health trajectories, sometimes more than genetic factors~\cite{ritchie2013our}. As a result, urban planners are increasingly focused on measuring and improving the walking environment~\cite{leyden2024walkable}. 
However, quantifying walkability is challenging because it depends not only on proximity to destinations~\cite{bokhari2016much}, but also on sidewalk conditions~\cite{Rowangould2019SidewalkQuality} and streetscape aesthetics~\cite{Tabatabaie2023SidewalksTreesShade}
, as well as population dynamics that reflect how people interact with their surroundings~\cite{diaz2019whose,resor2024simple}. Capturing these qualities often requires field audits or surveys, which are costly and labor-intensive, making it difficult to cover many neighborhoods or multiple cities and to repeat measurements regularly.

Advances in computer vision and the growing availability of urban imagery enable virtual audits of the built environment. Street view imagery, such as Google Street View, captures the pedestrian perspective and shows features such as sidewalks and greenery that is related with the walking experience~\cite{doiron2022predicting,kang2023assessment}. Satellite imagery captures the overhead perspective, revealing urban patterns such as land use and street connectivity~\cite{klemmer2025satclip,ouyang2024health}. However, prior studies typically rely on either street view or satellite imagery alone~\cite{doiron2022predicting,kang2023assessment,hosseini2021sidewalk}, which limits their ability to capture pedestrian conditions and the broader spatial context that jointly influence walkability.

Visual data (e.g., street view or satellite imagery) also have limitations alone. Imagery reveals physical structures but does not capture how people use or experience those spaces. Non-visual data sources, including search trends and environmental variables, provide behavioral signals~\cite{sun2024community}. The Population Dynamics Foundation Model (PDFM)~\cite{agarwal2024general} encodes these non-visual geospatial patterns into global spatial representations. However, PDFM lacks the visual information needed to describe the built environment.

To capture the physical and behavioral dimensions of walkability, we introduce WalkCLIP, a multimodal framework that integrates three complementary perspectives: street-level imagery for pedestrian infrastructure, satellite imagery for urban layout, and population dynamics data for behavioral and environmental context. Visual input represents the physical structure and appearance of urban spaces, while population dynamics representations summarize ZIP Code level aggregated activity patterns, local amenities, and environmental conditions that shape how neighborhoods are used.

Specifically, WalkCLIP learns walkability-aware vision–language representations by finetuning CLIP models on image-caption pairs generated with GPT-4o. WalkCLIP then applies a Spatially-Aware Feature Enhancement (SAFE) module that aggregates features from spatially adjacent locations to incorporate neighborhood context. The visual features produced by SAFE are fused with the Population Dynamics Foundation Model (PDFM) representations~\cite{agarwal2024general} to predict walkability scores. Evaluated on 4,660 geo-locations across Minneapolis and Saint Paul, WalkCLIP achieves an \(R^2\) of 0.887, outperforming PDFM, unimodal, and multimodal baselines in both predictive accuracy and spatial alignment.
\section{Related Work}
\subsection{Urban Walkability} 
Walkability refers to how easily residents can access destinations while experiencing a safe and comfortable walking environment~\cite{saelens2008built}. One of the most widely used tools for quantifying walkability is Walk Score, a commercial metric that rates neighborhoods based on the proximity of amenities such as grocery stores, schools, parks, and public transit~\cite{bokhari2016much}. In addition to Walk Score, urban planners have developed a range of walkability indices using GIS-based indicators, including residential density, intersection density, and access to amenities~\cite{frank2005linking}. For instance, the U.S. Environmental Protection Agency’s Walkability Index ranks neighborhoods by spatial accessibility using built environment variables~\cite{us2021national}. 

While these indices provide insights into access and connectivity, they exclude qualitative factors such as sidewalk conditions and streetscape appeal. As a result, traditional indices explain only part of the variation in walking behavior. This has led to growing interest in alternative data sources (e.g., urban imagery) and methods that can capture the pedestrian experience~\cite{kang2023assessment}.

\subsection{Visual Data for Urban Analysis}
To address the limitations of traditional walkability indices, recent work has explored the use of visual data to capture urban characteristics that influence the pedestrian experience. Street-level imagery, such as Google Street View, offers observations of pedestrian infrastructure and streetscape conditions, including sidewalks, crosswalks, greenery, building frontages, and street furniture~\cite{doiron2022predicting, kang2023assessment}. Complementing this, satellite imagery provides a top-down view of urban structure, capturing large-scale patterns such as street connectivity, land use diversity, and development density~\cite{klemmer2025satclip}. 

By applying computer vision techniques, researchers have used both modalities to conduct scalable “virtual audits” of the built environment. However, these prior work focuses on either street view or satellite imagery in isolation, limiting the ability to jointly capture physical features relevant to walkability.

\subsection{Behavioral Data for Urban Analysis}
While visual data capture the physical structure of urban environments, they do not reflect how spaces are used or experienced by people. Behavioral and environmental datasets, such as aggregated search trends, weather, and air quality, offer complementary signals that influence walkability. These sources capture patterns of human activity and environmental exposure that static imagery alone cannot reveal. Recent work has aimed to unify such data into structured spatial representations. Notably, the Population Dynamics Foundation Model (PDFM) encodes diverse geospatial indicators using a graph neural network to generate location-level representations~\cite{agarwal2024general}. PDFM has shown strong performance across a wide range of geospatial prediction tasks, including health, socio-economic, and environmental indicators. 

However, PDFM does not incorporate visual inputs and therefore cannot represent the physical features of the built environment. Integrating visual and non-visual data presents a promising approach to developing comprehensive urban models that capture both the appearance and usage of places.

\subsection{Multimodal Learning for Urban Analysis}
The success of Contrastive Language-Image Pre-training (CLIP)~\cite{radford2021learning} has inspired a wave of domain-specific adaptations that integrate visual encoders into geospatial modeling. RemoteCLIP~\cite{liu2024remoteclip} aligns satellite imagery with textual supervision to support land use and environmental analysis. SatCLIP~\cite{klemmer2025satclip} extends this idea by learning geographic representations from satellite imagery for a variety of downstream tasks. These works demonstrate the value of adapting vision-language models to spatial problems, especially when finetuned with domain-specific data. However, most approaches rely on a single type of visual input and do not address walkability.

We build on this line of work with WalkCLIP, a multimodal framework that integrates street-level imagery, satellite imagery, and behavioral context data to predict neighborhood walkability. Street-level imagery provides visual cues about pedestrian conditions, satellite imagery captures spatial layout, and population data reflects how areas are used. This combination enables accurate walkability prediction by linking physical appearance with population dynamics context.
\section{Method: WalkCLIP}
\begin{figure*}[!ht]
\includegraphics[width=\textwidth]{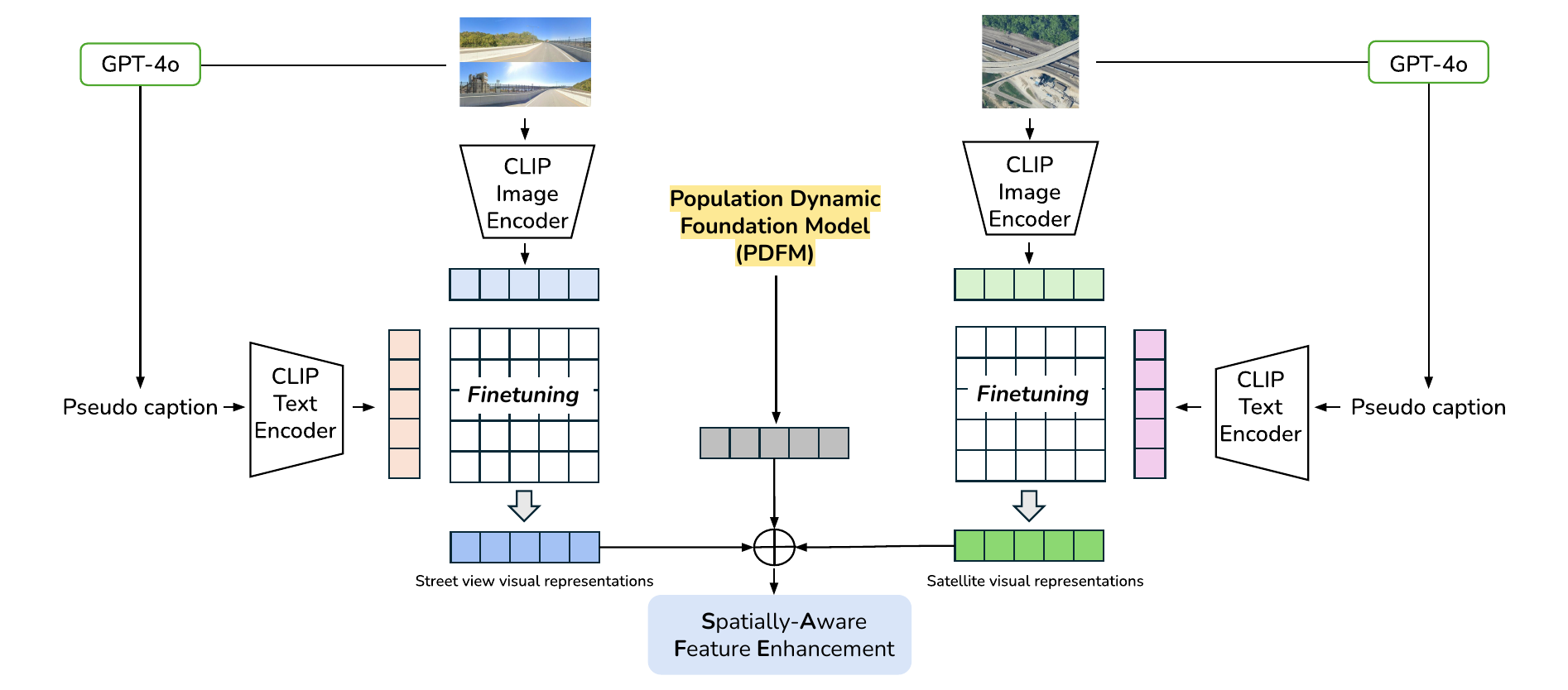}
\caption{The WalkCLIP framework with three stages. (1) Visual representations are learned by finetuning a CLIP transformer on satellite and street view images paired with GPT-4o-generated captions. (2) The Spatially-Aware Feature Enhancement (SAFE) module builds a proximity graph and aggregates features from nearby locations to produce spatially-aware representations. (3) Visual representations are fused with non-visual geospatial context from the Population Dynamics Foundation Model (PDFM)~\cite{agarwal2024general}. The final representations are then used to predict walkability scores.}
\label{fig:walkclip}
\end{figure*}

WalkCLIP is a multimodal contrastive learning approach for predicting neighborhood walkability that integrates visual data from satellite and street-level imagery with non-visual geospatial context such as population and behavioral patterns (Figure~\ref{fig:walkclip}). The framework comprises three components. (1) WalkCLIP learns walkability-aware vision–language representations that align visual inputs with corresponding textual descriptions, capturing pedestrian features from street-level views as well as broader urban form from satellite imagery. (2) Spatially-Aware Feature Enhancement (SAFE) module aggregates visual features from spatially adjacent neighborhoods to capture spatial dependencies. (3) Multimodal fusion stage combines visual and contextual signals into a unified representation used for walkability prediction.

\subsection{Stage 1: Walkability-Aware Vision-Language representations}
The goal of this stage is to learn visual representations of the built environment that emphasize features relevant to walkability. 
\paragraph{Why CLIP?}
Our intuition is that a generic visual backbone (e.g., supervised on ImageNet) tends to extract task-agnostic textures or object cues that are only loosely related to walkability. In contrast, CLIP aligns images with language, giving us a semantic handle to \emph{steer} the encoder toward walkability-relevant cues via caption supervision. By pairing each image with walkability-focused text, the model is encouraged to emphasize pedestrian infrastructure and urban form while discounting incidental content. 

Standard CLIP models are trained on web-scale image–text pairs ~\cite{radford2021learning} and do not include geographic information for urban form or pedestrian-level cues. To adapt the representations, we finetune a pretrained vision transformer encoder from CLIP~\cite{radford2021learning} on both satellite and street view imagery. Each image is paired with a descriptive caption generated by GPT-4o~\cite{manvi2023geollm,sahoo2024systematic}. 

\paragraph{Prompt design.}
We use template-based prompts for caption generation with GPT-4o; the same prompts are used across all experiments.

\noindent\textbf{Satellite prompt (structural, top-down).}
\begin{quote}\small
\texttt{Describe the walkability-related features you can infer from this satellite image with comma-separated phrases. Focus on block layout, street connectivity, parks/green space, and development density. Avoid details that are not visible from above.}
\end{quote}

\noindent\textbf{Street-view prompt (visible, ground-level).}
\begin{quote}\small
\texttt{Only name the walkability-related elements you actually see in this 4-view street-level composite, using comma-separated descriptive phrases. Do not infer unseen destinations; list visible items such as sidewalks, crosswalks, building frontages, street trees, lighting, bike lanes, and foot traffic.}
\end{quote}

The satellite prompts guide GPT-4o to describe structural features relevant to walkability, such as block layout, street connectivity, and green space. The street view prompts guide GPT-4o to list only visible pedestrian-level elements, including sidewalks, crosswalks, building frontages, trees, lighting, and foot traffic.

These captions serve as pseudo-labels in a contrastive learning setup. Given an image $x_i$ and caption $t_i$, the visual encoder $f_\theta(x_i)$ and text encoder $g_\phi(t_i)$ are optimized with the contrastive loss:
\begin{equation}
\mathcal{L}_{\text{CLIP}} = - \frac{1}{N} \sum_{i=1}^N 
\log \frac{\exp\big(\text{sim}(f_\theta(x_i), g_\phi(t_i))/\tau\big)}
{\sum_j \exp\big(\text{sim}(f_\theta(x_i), g_\phi(t_j))/\tau\big)},
\end{equation}
where $\text{sim}(\cdot,\cdot)$ is cosine similarity and $\tau$ is a learnable temperature.

We finetune two separate pretrained CLIP models to learn visual representations: one adapted to satellite imagery and the other to street-view imagery, each paired with GPT-4o–generated captions. This yields two modality-specific representation spaces. For each location, we obtain a satellite representation $f_{\text{sat}} \in \mathbb{R}^d$ and a street view representation $f_{\text{street}} \in \mathbb{R}^d$. Although the two modalities are co-registered to the same site, their representations provide complementary perspectives on walkability.

\subsection{Stage 2: Spatially-Aware Feature Enhancement (SAFE)}
\begin{figure}[!h]
    \centering
    \includegraphics[width=\linewidth]{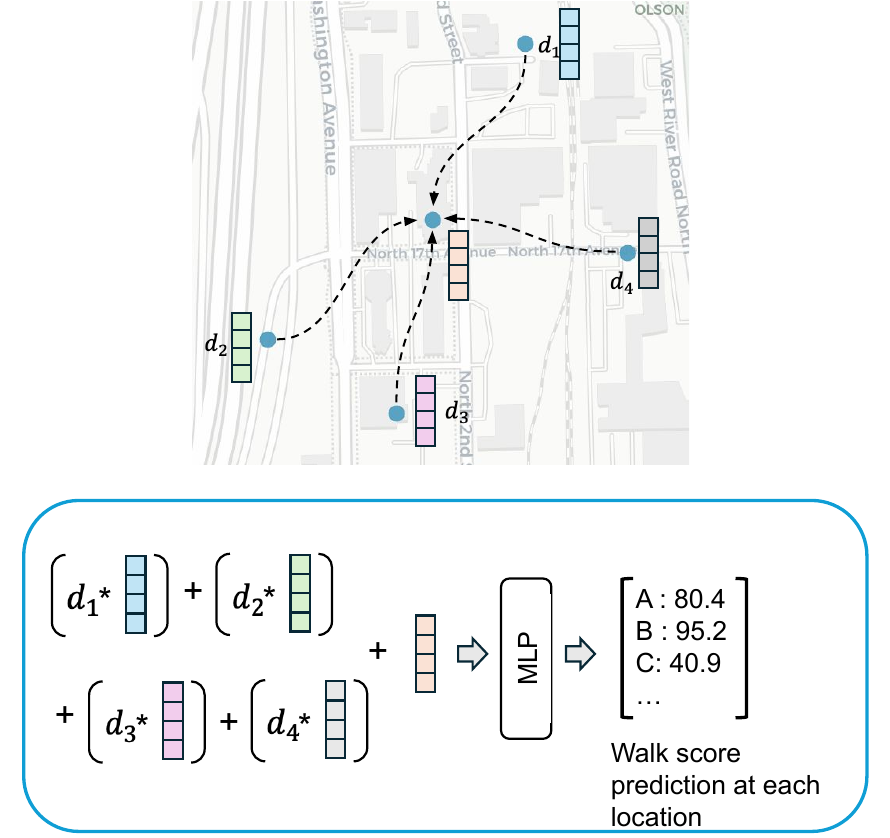}
    \caption{Overview of the Spatially-Aware Feature Enhancement (SAFE) module. Each location aggregates visual representations from nearby neighbors with distance-based weights. The combined representation is passed through a multilayer perceptron (MLP) to predict walkability scores.}
    \label{safe}
\end{figure}

Walkability is shaped not only by conditions at a specific location but also by the surrounding neighborhood~\cite{singh2016factors}. For example, a residential block may feel walkable if it is near parks, shops, or a well-connected street network. Since walkability depends on both local and neighborhood context, we introduce the Spatially-Aware Feature Enhancement (SAFE) module to incorporate visual information from nearby areas into each representation (\autoref{safe}).

SAFE operates on the set of geo-located points from our dataset. For each point $i$
, its set of neighbors $\mathcal{N}(i)$, is defined as all other points whose straight-line coordinate distance in (lat, lon) space from $i$ is less than 0.01°. SAFE then updates the representation for point $i$ by performing a weighted aggregation of features from itself and its neighbors:
\begin{equation}
f'_i = \frac{1}{Z_i} \left( w_{ii} f_i + \sum_{j \in \mathcal{N}(i)} w_{ij} f_j \right),
\end{equation}
The weights $w_{ij}$ are determined using an Inverse Distance Weighting (IDW) scheme with a power parameter of 1:
\begin{equation}
w_{ij} = \frac{1}{d(i,j) + \epsilon}.
\end{equation}

Here, $d(i,j)$ is the geographic distance between $i$ and $j$, $\epsilon$ is a small constant to avoid division by zero, and $Z_i = \sum_{j \in \mathcal{N}(i)} w_{ij} + w_{ii}$ is a normalization factor. This ensures that closer neighbors contribute more strongly than distant ones.

By giving higher weight to closer neighbors, SAFE encodes spatial dependence into the representations. We apply SAFE separately to the satellite and street view features, producing weighted representations for walkability prediction.

\subsection{Stage 3: Multimodal Fusion and Prediction}
Visual features capture the physical structure of a location, but walkability is also influenced by population behavioral and environmental conditions~\cite{baobeid2021walkability}. For example, two visually similar areas may differ in walkability if one has higher pedestrian activity or better air quality than the other region. To incorporate these non-visual factors, we use the Population Dynamics Foundation Model (PDFM), which encodes patterns of human activity, like web search trends and point-of-interests, and various environmental conditions, including weather or air quality, by graph-based approach~\cite{agarwal2024general}.

For each location, we retrieve the corresponding ZIP Code level PDFM representation $f_{\text{pdfm}}$. We then fuse this representation with the SAFE-processed visual features from satellite and street view imagery. The fused representation is constructed as
\begin{equation}
h_i = [f_{\text{sat}}; f_{\text{street}}; f_{\text{pdfm}}],
\end{equation}
where $[\cdot]$ denotes vector concatenation. The walkability score is predicted using an MLP:
\begin{equation}
\hat{y}_i = \text{MLP}(h_i).
\end{equation}
The MLP is trained with mean squared error (MSE) loss:
\begin{equation}
\mathcal{L}_{\text{MSE}} = \frac{1}{N} \sum_{i=1}^N (y_i - \hat{y}_i)^2,
\end{equation}
where $y_i$ is the reference value and $\hat{y}_i$ is the prediction.
\section{Experiment Setup}
\begin{figure}[h!]
\centering
\includegraphics[width=\linewidth]{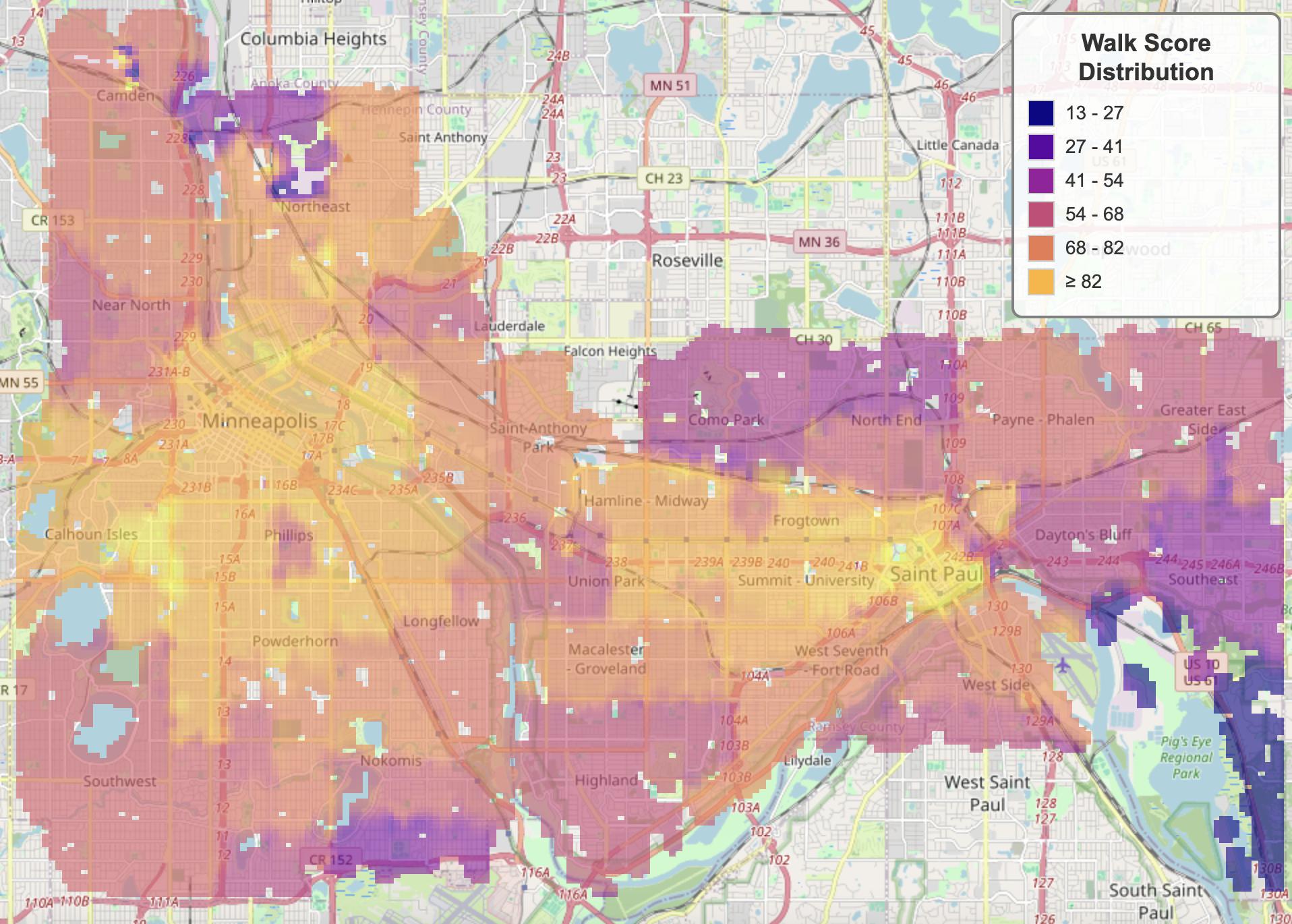}
\caption{Walk Score distribution across Minneapolis–Saint Paul. Each cell represents a neighborhood-level area colored by its Walk Score value. These scores reflect the walkability of the built environment based on factors such as proximity to amenities, street connectivity, and pedestrian infrastructure. Higher scores (yellow) indicate highly walkable areas, while lower scores (dark blue) correspond to less walkable regions.}
\label{fig:map}
\end{figure}

\subsection{Datasets}
Our study is based on 4,660 geo-located points sampled from street intersections across Minneapolis and Saint Paul, Minnesota in United States. We retrieved the complete street networks for Minneapolis and Saint Paul from OpenStreetMap using OSMnx. This yielded 96,164 intersections in Minneapolis and 60,216 in Saint Paul. From each city, we drew a simple random sample of 2,500 intersections, for 5,000 candidate locations in total. After excluding locations without valid street view coverage or with imagery quality issues, 4,660 locations remained with both street view and satellite imagery, these 4,660 locations constitute our study sample.

\paragraph{Pretraining Data}
To capture urban structure, we obtained satellite imagery from Google Static Maps, Google aggregates imagery from commercial and governmental providers. The imagery is sourced from Maxar’s WorldView series satellites (e.g., WorldView-3 and WorldView-4), which provide a spatial resolution of approximately 0.3 meters per pixel, allowing for detailed visualization of streets, building footprints, and other fine-scale urban features. 

The imagery was centered on each target coordinate, then cropped and resized to $256\times$256 pixels for consistency. These views represent spatial features such as street layout, building footprints, and green space.

To capture pedestrian environments, we collected street-level imagery from Google Street View. For each location, we retrieved the nearest available panorama and extracted four directional views at bearings of 90°, 180°, 270°, and 360°, resized to 224$\times$224 pixels~\cite{li2022measuring}. These views provide pedestrian-level context, including sidewalks, crosswalks, building facades, trees, and street furniture.  

To capture behavioral and environmental context, we used 128-dimensional representations from the Population Dynamics Foundation Model (PDFM)~\cite{agarwal2024general} for each ZIP Code.

\paragraph{Why this dataset?}
Although Walk Score is largely driven by proximity to amenities, we adopt it here as a principled, reproducible, and widely‑queried baseline for preliminary evaluation. Its standardized scale and broad coverage make it well suited for cross‑model comparison and ablation studies at the city scale. We emphasize that Walk Score is used as a proxy and a first‑step benchmark rather than an authoritative ground truth: it omits important experiential and infrastructure quality dimensions (e.g., sidewalk condition, crossing safety, lighting, and temporal variations). These limitations motivate our multimodal design and qualitative analyses, future work will augment evaluation with human audits, and travel‑survey data to better capture experiential walkability.

To the best of our knowledge, no existing dataset combines co‑located street view imagery, satellite imagery, ZIP‑level population dynamics representations, and a standardized walkability score at this scale. We therefore constructed a new multimodal dataset to enable systematic development, benchmarking, and ablation of models like WalkCLIP that seek to integrate visual and behavioral perspectives on walkability.

\paragraph{Downstream Data}
We use Walk Score\footnote{\url{https://www.walkscore.com/}} as the supervised signal of walkability in our regression task. Walk Score assigns values from 0 (least walkable) to 100 (most walkable). Our dataset includes 4,660 unique geolocated points sampled at street intersections in Minneapolis (2,525) and Saint Paul (2,135), United States. These scores serve as target values for training and evaluating the walkability prediction models.

\subsection{Baselines}
We compare WalkCLIP to five baselines including both unimodal and multimodal models:
\begin{itemize}
\item \textbf{StreetCLIP~\cite{haas2023learning}}: trained a transformer encoder using CLIP-style contrastive learning on street view images paired with GPT-4o-generated captions.
\item \textbf{SatCLIP~\cite{klemmer2025satclip}}: trained a transformer encoder using CLIP-style contrastive learning on satellite images paired with GPT-4o-generated captions.
\item \textbf{Population Dynamics Foundation Model (PDFM)~\cite{agarwal2024general}}: generates ZIP Code level representations based on aggregated behavioral and environmental signals. PDFM is a baseline for purely non-visual geospatial context.
\item \textbf{VisionCLIP (Street view + Satellite)}: combines street view and satellite representations from SatCLIP and StreetCLIP using late fusion via concatenation. No non-visual features are used.
\item \textbf{VisionCLIPv2 (VisionCLIP + PDFM)}: combines visionCLIP with PDFM representations using late fusion via concatenation. This variant excludes the SAFE module.
\end{itemize}

\subsection{Evaluation Method \& Metrics}
We evaluate model performance using a two-stage protocol that combines a fixed hold-out test set with cross-validation for hyperparameter tuning. First, we apply a \texttt{GroupShuffleSplit} to partition the full dataset, setting aside 15\% of the data as a test set. This set remains completely unseen during training and validation. The remaining 85\% is used for training and model selection. On the training and validation portion, we perform 5-fold cross-validation using a \texttt{StratifiedGroupKFold}. Stratification is based on quartile bins of the target variable (Walk Score) to ensure a balanced distribution of walkability scores across folds. Grouping is enforced by location ID to prevent augmented views of the same location from leaking across folds.

The final model is evaluated on the held-out test set. We report \(R^2\) and RMSE, which are standard metrics for spatial prediction~\cite{jean2016combining,xi2022beyond}. To assess spatial alignment, we compute the sliced Wasserstein distance (SWD)~\cite{rabin2011wasserstein}, an efficient approximation of the Wasserstein (Earth Mover’s) distance~\cite{rubner2000earth}. SWD compares distributions by projecting them onto multiple one-dimensional directions and averaging the resulting Wasserstein distances. SWD is computed over geolocated prediction-target pairs, allowing the model to capture value similarity and the spatial proximity of errors.

\subsection{Implementation Details}
All CLIP models were finetuned with the Adam optimizer for up to 10 epochs on the image-caption dataset generated via GPT-4o. To improve robustness, we applied data augmentation to visual inputs using the Albumentations library~\cite{buslaev2020albumentations}, including \textit{HorizontalFlip}, \textit{VerticalFlip}, \textit{ColorJitter}, and \textit{RandomRotate90}. Augmented images were assigned the same location ID as their originals to preserve grouping during cross-validation.

For the downstream regression task, we trained a multi-layer perceptron (MLP) on the final feature representations. To find the optimal settings, we used a 5-fold stratified group cross-validation with a grid search over hyperparameters, including learning rate (1e-3, 1e-4), dropout rate (0.3, 0.5), and weight decay (1e-3, 1e-4). The MLP was optimized using the AdamW optimizer and included dropout layers for regularization. The best-performing hyperparameter combination was then used to train the final model on the entire training set for 150 epochs.
\section{Results \& Discussion}
The experimental results are summarized in Table~\ref{tab:results}. We compare WalkCLIP against several baselines and ablated variants to assess the contribution of each modality and component. The results reveal a consistent trend: predictive performance improves as additional modalities and spatial context are incorporated into the model.

\begin{table}[!ht]
    \centering
    \caption{Performance comparison and ablation study. Results show that predictive accuracy improves as additional modalities and spatial structure are incorporated. +SAFE indicates the use of the Spatially-Aware Feature Enhancement module.}
    \label{tab:results}
    \resizebox{\columnwidth}{!}{%
    \begin{tabular}{@{}llccc@{}}
        \toprule
        & \textbf{Methods} & \textbf{Test \(R^2\)} & \textbf{Test RMSE} & \textbf{SWD} \\
        \midrule
        (1) & StreetCLIP & 0.350 & 11.620 & 4.860 \\
        (2) & SatCLIP & 0.538 & 9.795 & 3.442 \\
        (3) & PDFM & 0.616 & 8.937 & 3.170 \\
        \midrule
        (4) & VisionCLIP (Street view + Satellite) & 0.597 & 9.154 & 2.936 \\
        (5) & VisionCLIPv2 (VisionCLIP + PDFM)& 0.782 & 6.724 & 1.933 \\
        (6) & \textbf{WalkCLIP (VisionCLIPv2 + SAFE)} & \textbf{0.887} & \textbf{4.838} & \textbf{1.100} \\
        \bottomrule
    \end{tabular}%
    }
\end{table}

\subsection{Quantitative Analysis}
\paragraph{Single-Modality Performance.} Among the single-modality baselines, the model using only PDFM representations achieves the highest performance ($R^2 = 0.616$), showing that behavioral and environmental context provides a strong predictive signal for walkability. The SatCLIP model ($R^2 = 0.538$) outperforms StreetCLIP ($R^2 = 0.350$), suggesting that satellite imagery captures broader spatial patterns that are beneficial than ground-level views alone.

\paragraph{Multimodal Integration} Combining satellite and street view inputs in the VisionCLIP model improves performance to $R^2 = 0.597$, exceeding either visual modality alone. This indicates that the two perspectives offer complementary information relevant to walkability. Adding PDFM representations to the visual features (VisionCLIP + PDFM) further boosts performance to $R^2 = 0.782$, highlighting the value of integrating non-visual context with image-based features.

\paragraph{Effectiveness of SAFE}
Our model, WalkCLIP, which incorporates all modalities along with the Spatially-Aware Feature Enhancement (SAFE) module, achieves the highest performance ($R^2 = 0.887$). This improvement indicates that leveraging spatial context from nearby regions allows the model to make more accurate predictions aligned with ground-truth walkability.

\paragraph{Spatial Consistency} 
The sliced Wasserstein distance (SWD) provides further insight into the spatial quality of predictions. Single-modality models exhibit high SWD values (e.g., StreetCLIP: 4.860, SatCLIP: 3.442, PDFM: 3.170), indicating less spatial alignment between predictions and ground truth. The Vision + PDFM model lowers the SWD to 1.933, while WalkCLIP achieves the lowest SWD of 1.100. This result suggests that WalkCLIP not only improves accuracy but also produces spatially consistent predictions.

\subsection{Model Interpretability}
\begin{figure}[!ht]
\centering

\begin{tabular}{@{}cc@{}}
  \includegraphics[width=0.48\columnwidth]{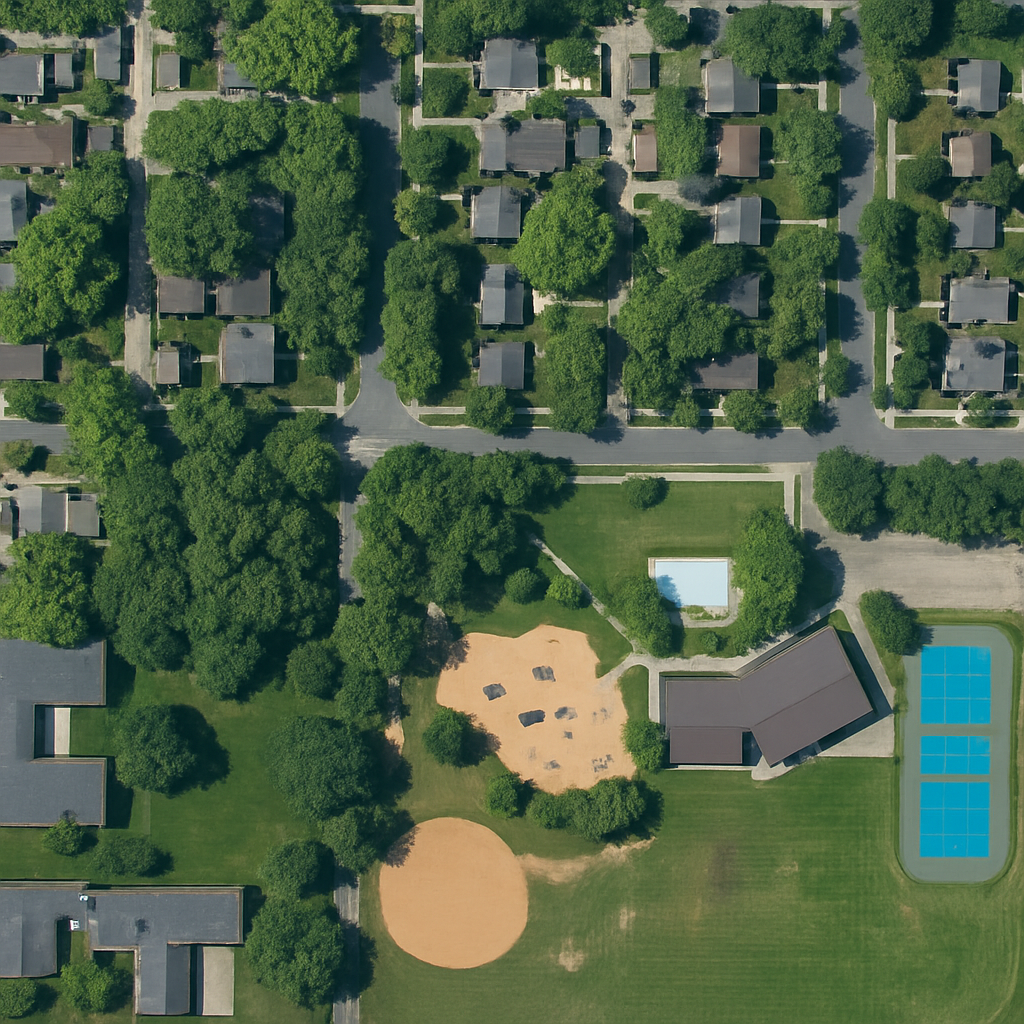} &
  \includegraphics[width=0.48\columnwidth]{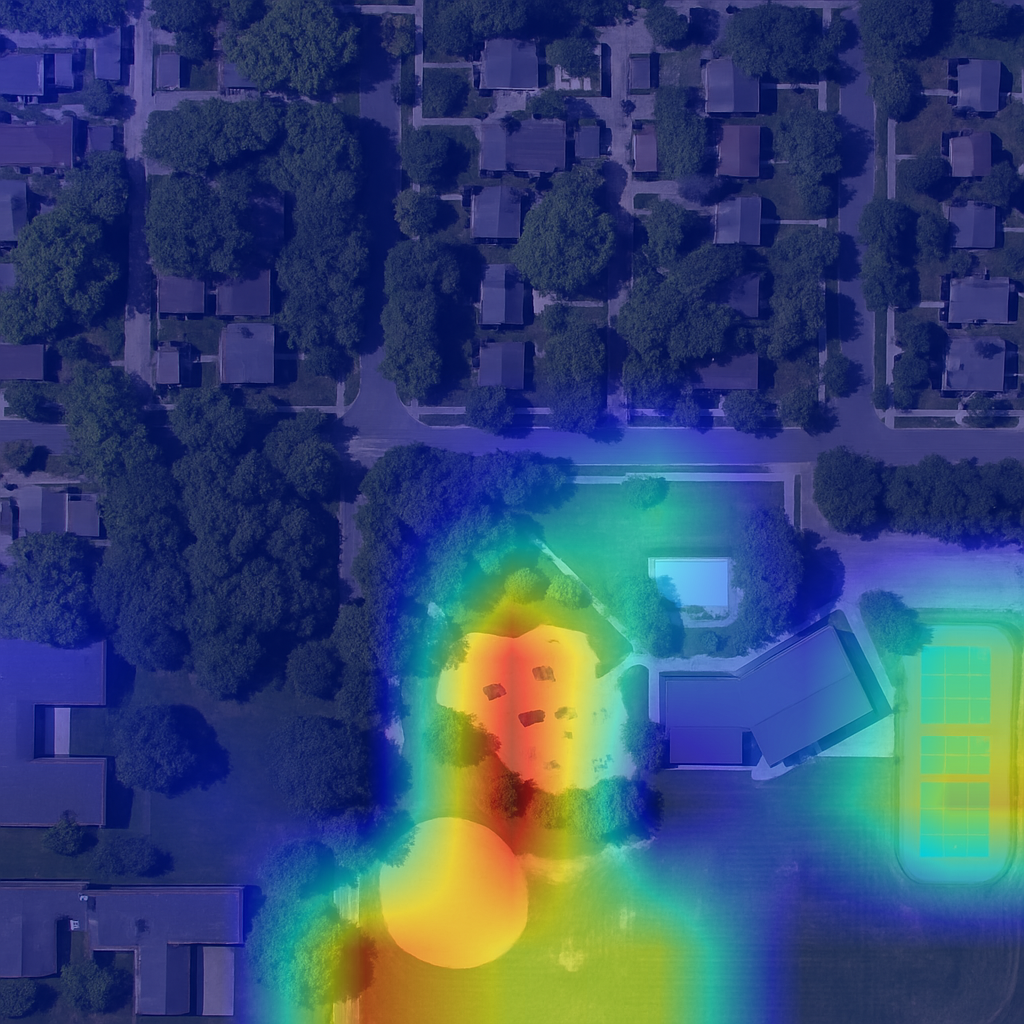}
\end{tabular}

\vspace{0.5em}
\parbox{\columnwidth}{
  \footnotesize
  \textbf{GPT-4o Generated Caption:} Residential streets, tree-lined sidewalks, nearby parks, recreational facilities, tennis courts, walking paths, proximity to community amenities.
}

\vspace{0.75em}
\centerline{(a) Satellite Image (Bird's Eye)}

\vspace{1.5em}

\begin{tabular}{@{}cc@{}}
  \includegraphics[width=0.48\columnwidth]{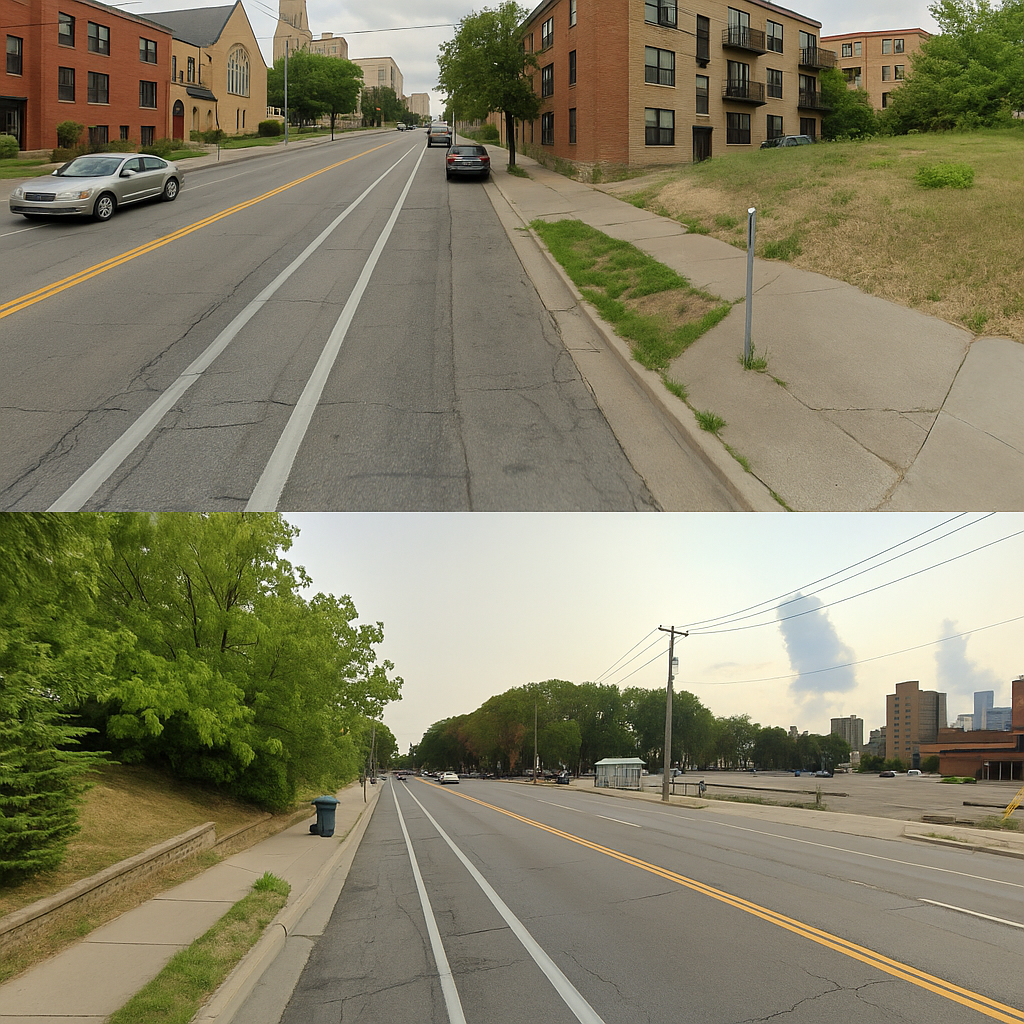} &
  \includegraphics[width=0.48\columnwidth]{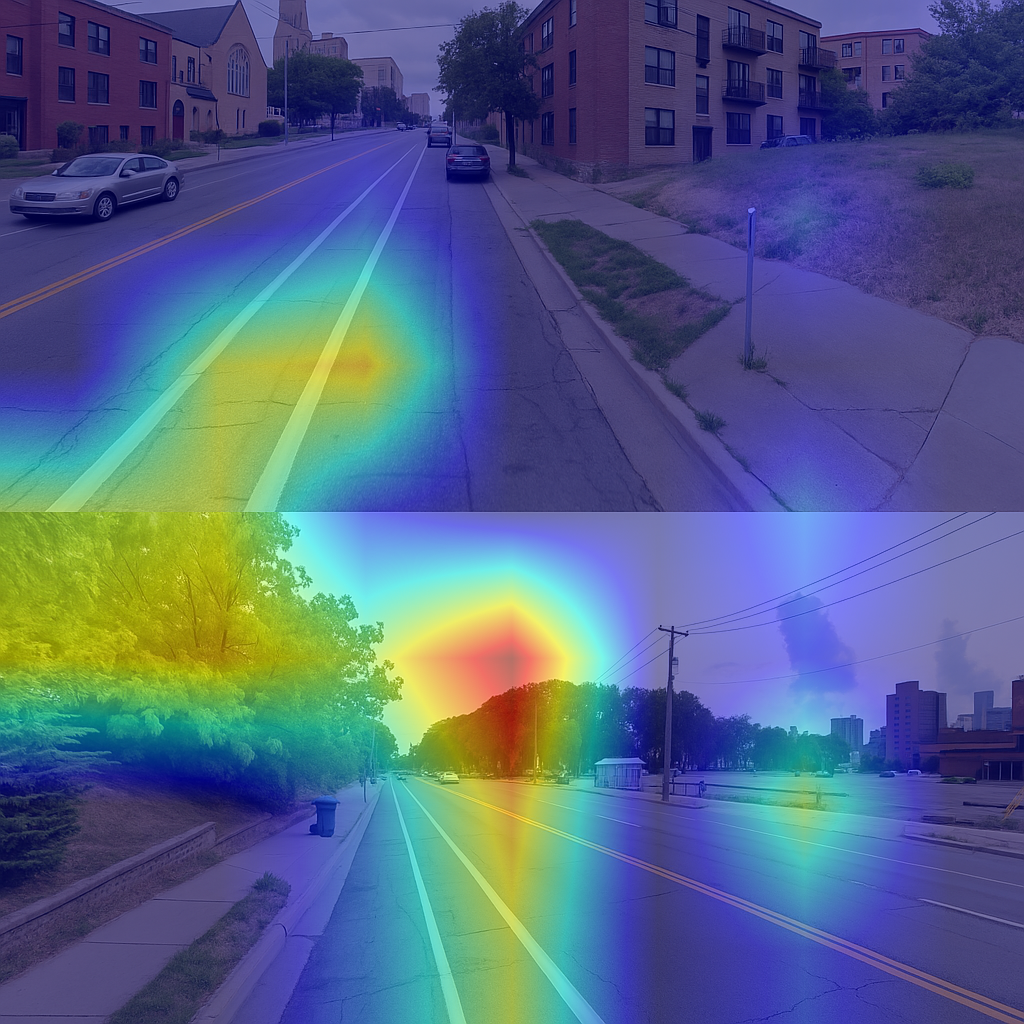}
\end{tabular}

\vspace{0.5em}
\parbox{\columnwidth}{
  \footnotesize
  \textbf{GPT-4o Generated Caption:} Bike lanes, sidewalks, residential buildings, trees, parked cars, power lines, streetlights, open green space, nearby commercial buildings.
}

\vspace{0.75em}
\centerline{(b) Street view Image (Ground Level)}

\caption{Attention heatmaps from WalkCLIP highlighting visual cues associated with walkability. (a) In satellite imagery, the model focuses on residential street layouts, sidewalks, and community amenities such as parks, playgrounds, and recreational facilities. (b) In street view imagery, high attension score are on sidewalks, bike lanes, and tree cover.}
\label{fig:interpretability}
\end{figure}

In walkability prediction, it is important to identify which visual elements the model relies on~\cite{dwivedi2023explainable} and to check whether they align with human-understandable features such as sidewalks, parks, or access to public transit. To examine the decision-making process of WalkCLIP, we generate attention heatmaps~\cite{samek2017explainable} (see Figure~\ref{fig:interpretability}).

From the satellite perspective (Figure~\ref{fig:interpretability}a), the model’s attention concentrates on recreational amenities such as the playground, baseball field, and tennis courts. This indicates that the satellite encoder identifies land-use patterns associated with public space and community activity, which are key components of walkability. From the street view perspective (Figure~\ref{fig:interpretability}b), the encoder focuses on ground-level infrastructure, including bike lanes, sidewalks, transit stops, and residential buildings. These cues are linked to pedestrian accessibility, transportation options, and the walking experience.

Together, these visualizations show that the modality-specific encoders learn to extract distinct yet complementary evidence of walkability. The satellite encoder captures land-use and amenity patterns, while the street view encoder detects ground-level infrastructure. This division of focus supports the design of our multimodal framework and highlights its ability to learn a comprehensive representation of the urban environment.

\section{Case Study}
\begin{figure}[!ht]
  \centering
  \begin{tabular}{cc}
    \includegraphics[height=3.2cm, width=\linewidth, keepaspectratio]{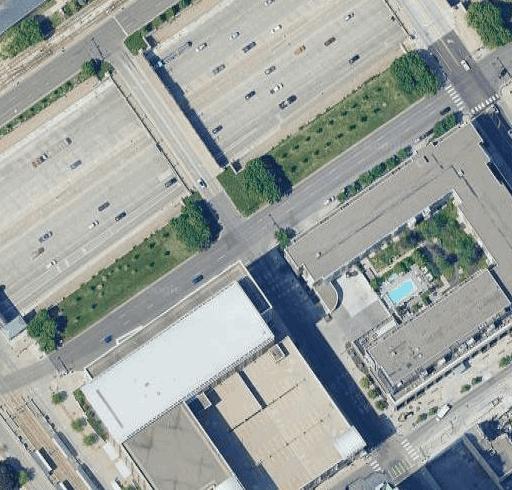} &
    \includegraphics[height=3.2cm, width=\linewidth, keepaspectratio]{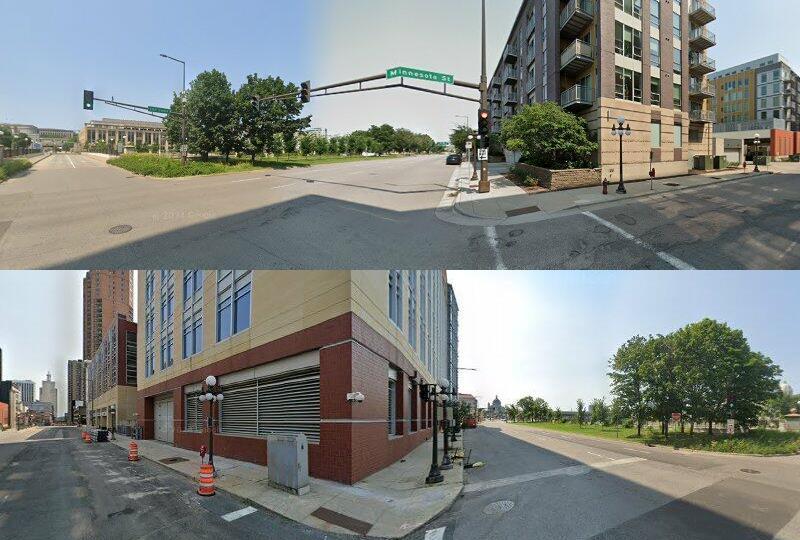} \\
    \multicolumn{2}{c}{(a) Case 1: Redeveloped downtown core of St. Paul} \\[1ex]
    \includegraphics[height=3.2cm, width=\linewidth, keepaspectratio]{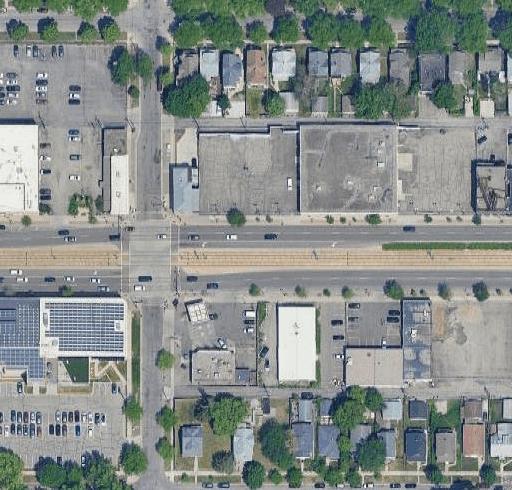} &
    \includegraphics[height=3.2cm, width=\linewidth, keepaspectratio]{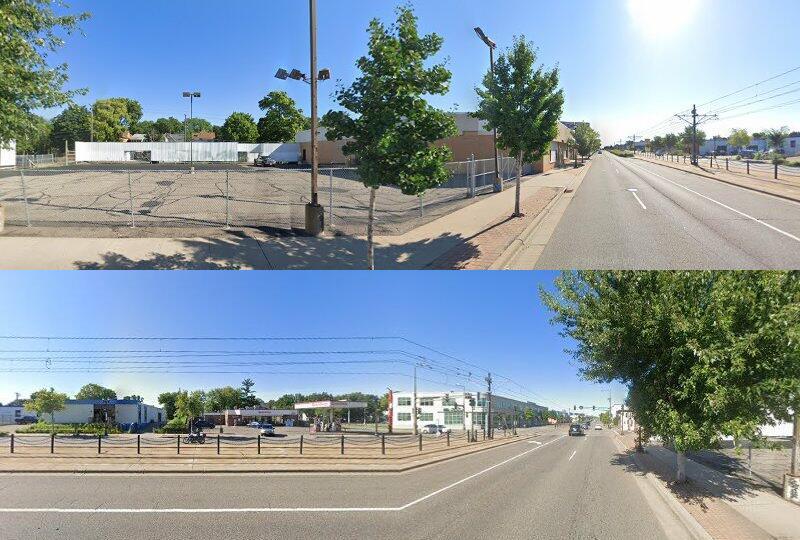} \\
    \multicolumn{2}{c}{(b) Case 2: Major commercial area in St. Paul} \\[1ex]
    \includegraphics[height=3.2cm, width=\linewidth, keepaspectratio]{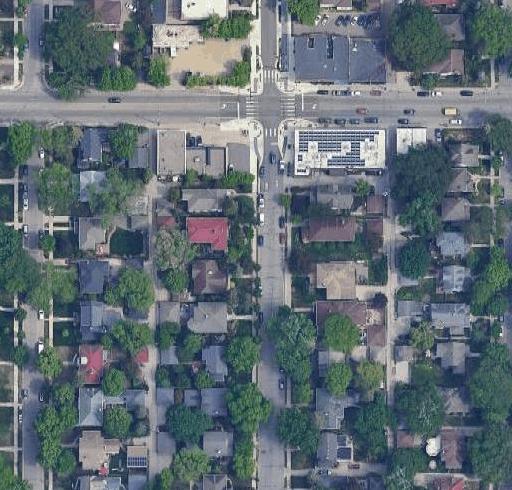} &
    \includegraphics[height=3.2cm, width=\linewidth, keepaspectratio]{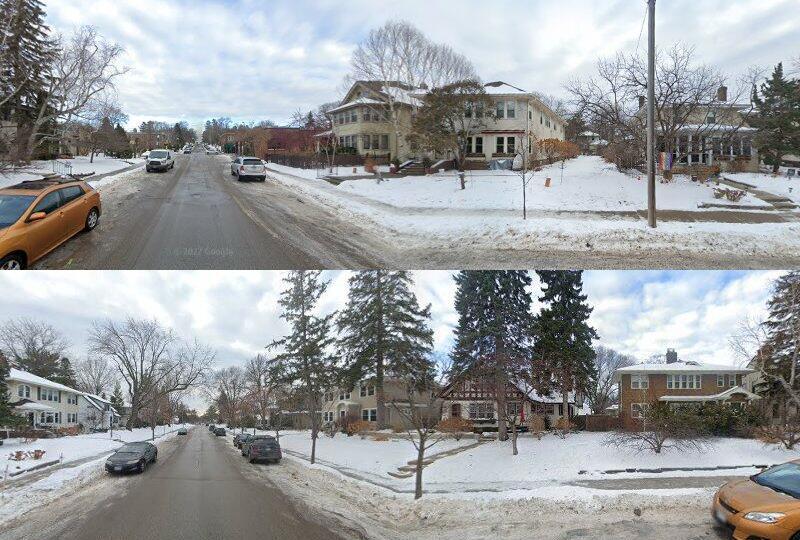} \\
    \multicolumn{2}{c}{(c) Case 3: Residential neighborhood in Southwest Minneapolis} \\
  \end{tabular}
  \caption{Case study examples showing paired satellite (left) and street view (right) imagery.}
  \label{fig:case1}
\end{figure}

\subsection{Modality-Specific Patterns in Walkability Predictions}
To illustrate how different modalities contribute to walkability prediction, we analyze three unseen locations (\autoref{fig:case1}). We compare a vision-only CLIP model that fuses satellite and street view representations, a context-only model based on population and environmental signals (i.e., PDFM), and the combined WalkCLIP model. These case studies highlight when each modality is informative, and how their combination provides balanced predictions.

\paragraph{Case 1 (44.95°N, 93.10°W): Redeveloped downtown core of St. Paul}
This location lies in the redeveloped downtown core of St. Paul, characterized by high-density housing, mixed land use, and upgraded pedestrian infrastructure. The satellite image (Figure~\ref{fig:case1}a) shows compact urban blocks, while street view imagery highlights sidewalks, signalized intersections, and new building frontages. VisionCLIP, which relies only on visual signals, assigns a very high score (92.4), strongly influenced by these visible cues of urban form. WalkCLIP predicts a slightly lower score (87.6; Walk Score: 90.0), as contextual information tempers the purely visual estimate and brings it closer to the proximity-based reference. By contrast, the context-only PDFM model assigns a much lower score (52.1), likely because population and activity signals lag behind recent redevelopment. This case illustrates how vision captures newly built pedestrian infrastructure, context-only inputs underrepresent rapid change, and WalkCLIP achieves balanced predictions.

\paragraph{Case 2 (44.96°N, 93.19°W): Major commercial area in St. Paul}
This location lies along University Avenue, a major commercial area in St. Paul with light rail service and frequent public activity. The satellite image (Figure~\ref{fig:case1}b) highlights large blocks and surface parking, while street view imagery shows wide roads, fencing, and sparse greenery, giving the area a car-oriented appearance. VisionCLIP, which relies solely on visual inputs, assigns a high score (85.7), emphasizing corridor density and development visible in imagery. WalkCLIP predicts a slightly lower score (80.1; Walk Score: 65.0), balancing visual impressions with contextual information. By contrast, the context-only PDFM model assigns a more moderate score (70.4), capturing transit accessibility and population activity not apparent from imagery. This case illustrates how contextual inputs can temper visually optimistic predictions.

\paragraph{Case 3 (44.91°N, 93.29°W): Residential neighborhood in Southwest Minneapolis}
This location is a residential neighborhood in Southwest Minneapolis characterized by higher-density housing, tree-lined streets, and regular pedestrian activity. The satellite image (Figure~\ref{fig:case1}c) shows a connected street grid with greenery, while street view imagery depicts continuous sidewalks, short block lengths, and closely spaced homes that support everyday walking. VisionCLIP predicts a high score (83.6), emphasizing the visible pedestrian infrastructure and compact street network. WalkCLIP and the context-only PDFM model also predict high scores (WalkCLIP: 82.0; PDFM: 79.5; Walk Score: 80.0), resulting in strong alignment across modalities. This case illustrates that when both visual and contextual cues indicate pedestrian accessibility, all models converge on consistent high predictions.

\subsection{Different Perspectives on Walkability}
\begin{figure}[!ht]
  \centering
  \begin{tabular}{cc}
    \includegraphics[height=3.2cm, width=\linewidth, keepaspectratio]{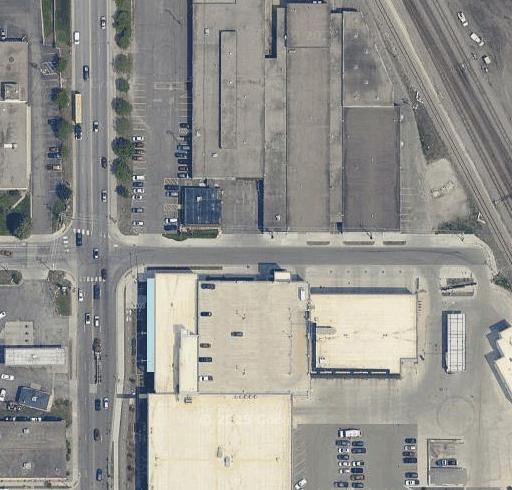} &
    \includegraphics[height=3.2cm, width=\linewidth, keepaspectratio]{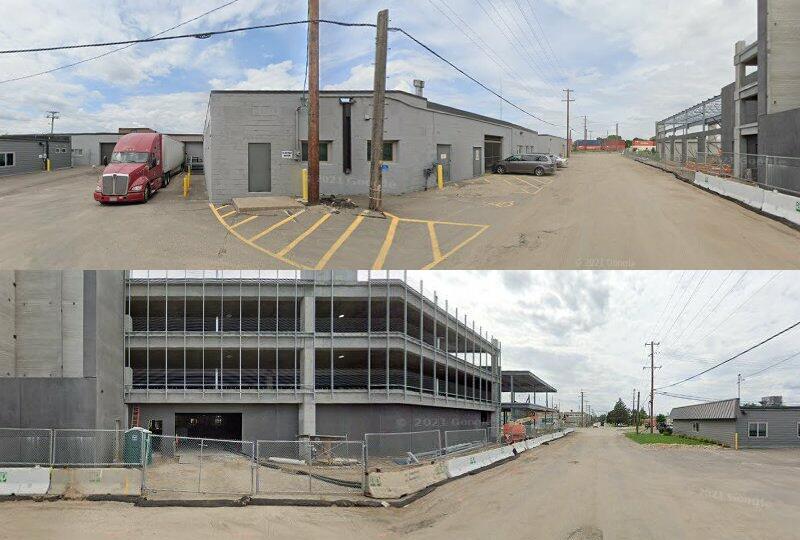} \\
    \multicolumn{2}{c}{(a) Case 1: Industrial Zone} \\[1ex]
    \includegraphics[height=3.2cm, width=\linewidth, keepaspectratio]{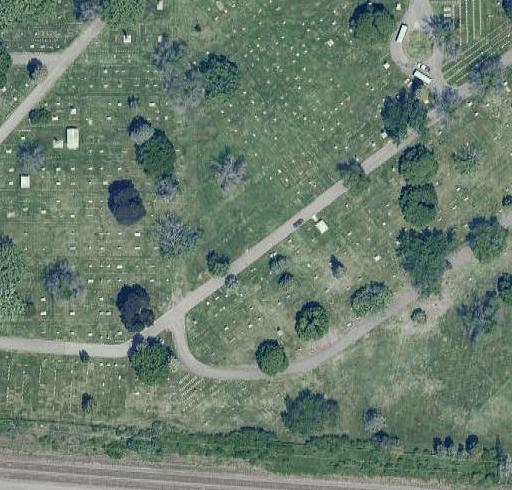} &
    \includegraphics[height=3.2cm, width=\linewidth, keepaspectratio]{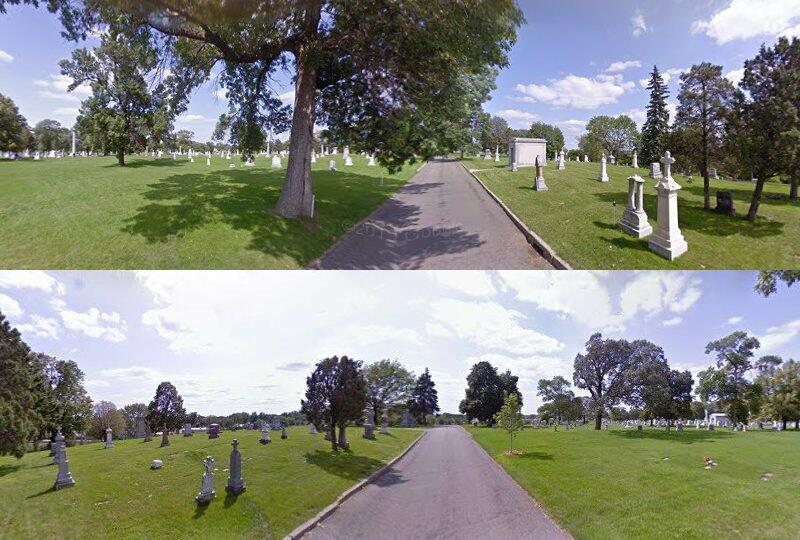} \\
    \multicolumn{2}{c}{(b) Case 2: Cemetery} \\[1ex]
    \includegraphics[height=3.2cm, width=\linewidth, keepaspectratio]{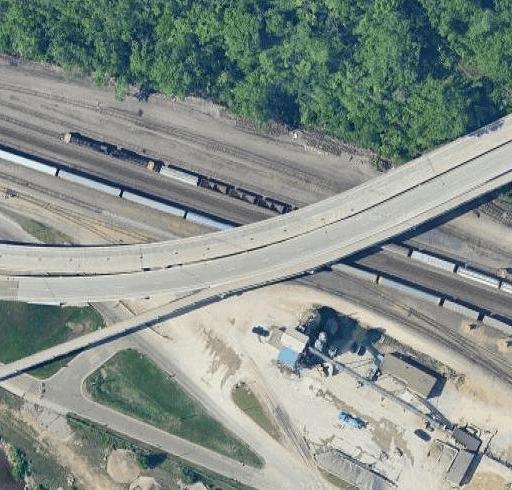} &
    \includegraphics[height=3.2cm, width=\linewidth, keepaspectratio]{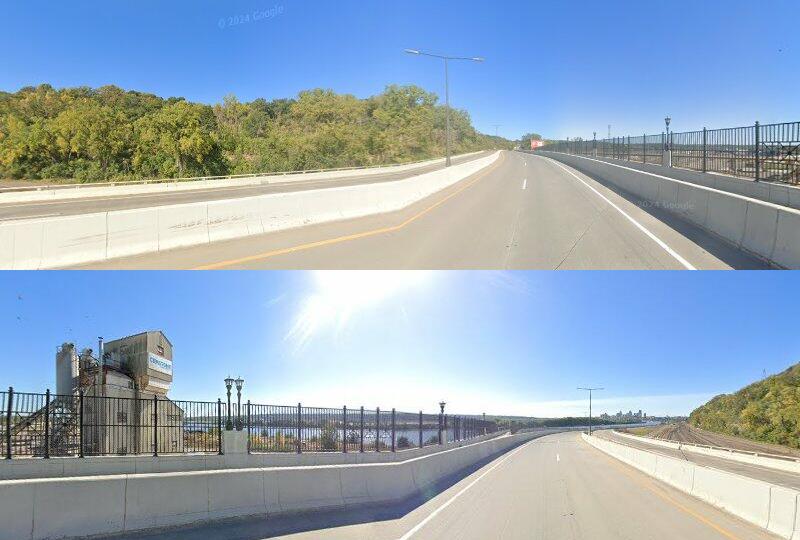} \\
    \multicolumn{2}{c}{(c) Case 3: Highway Bridge} \\
  \end{tabular}
  \caption{Examples where WalkCLIP predictions and Walk Score emphasize different aspects of walkability. (a) Industrial area where WalkCLIP assigns a higher score based on dense building footprints, while Walk Score reflects limited proximity to amenities. (b) Cemetery adjacent to a commercial district, where WalkCLIP predicts a lower score given the lack of functional pedestrian infrastructure, while Walk Score reflects close proximity to nearby destinations. (c) Highway bridge where WalkCLIP captures the presence of a fenced pedestrian path, while Walk Score emphasizes the absence of accessible amenities in the surrounding area.}
  \label{fig:disagreement}
\end{figure}

Walk Score primarily measures proximity to amenities, it can overlook the qualitative aspects of the pedestrian environment that visual and behavioral data are able to capture. For example, a high density of destinations does not guarantee the presence of safe, comfortable, or aesthetically pleasing infrastructure, such as sidewalks, crosswalks, and greenery. This section explores cases where WalkCLIP’s predictions diverge from Walk Score. These disagreements should not be interpreted as model errors; rather, they highlight the different dimensions of walkability each method emphasizes. By analyzing these specific instances, we aim to illustrate how a multimodal approach provides a holistic assessment by capturing the qualitative, experiential aspects of walkability that proximity-based scores alone cannot. (Figure~\ref{fig:disagreement}).

\paragraph{Case 1 (45.01°N, 93.26°W): Industrial Zone}
This location lies within an industrial zone and has a Walk Score of 41, reflecting limited accessibility to nearby amenities. The satellite view (Figure~\ref{fig:disagreement}a) shows large warehouses, expansive surface parking, and wide arterial roads. Street-level imagery highlights truck access points, construction activity, and the lack of sidewalks or crosswalks. WalkCLIP assigns a higher score of 73, likely because it interprets the concentration of large building footprints as a signal of urban density. In this case, Walk Score provides the more reliable assessment, since the dense industrial area does not translate into safe or functional pedestrian conditions.

\paragraph{Case 2 (44.96°N, 93.13°W): Cemetery}
This location lies within a cemetery but has a Walk Score of 82, reflecting close proximity to a nearby commercial district. The satellite image (Figure~\ref{fig:disagreement}b) shows a large, green parcel separated from the surrounding street grid. Street view imagery reveals orderly, well-maintained grounds with narrow internal roads but no sidewalks or crosswalks designed for everyday connectivity. WalkCLIP predicts a lower score of 53. Although it identifies the park-like setting, the reduced score reflects the absence of functional pedestrian infrastructure. Both perspectives are reasonable: Walk Score emphasizes proximity to neighborhood destinations, while WalkCLIP reflects the local visual environment, underscoring the nuance of spaces that appear green and structured but are not functionally walkable.

\paragraph{Case 3 (44.94°N, 93.05°W): Highway Bridge}
This location is on a highway bridge and has a Walk Score of 13, reflecting very limited proximity to everyday destinations. The satellite image (Figure~\ref{fig:disagreement}c) shows the bridge spanning rail lines and industrial land, with no nearby services. Street view imagery highlights a multi-lane highway but also reveals a fenced pedestrian path. WalkCLIP predicts a somewhat higher score of 41. Although it recognizes the presence of this dedicated path, the lower overall score reflects the surrounding car-oriented environment and lack of connected pedestrian infrastructure. Both perspectives make sense: Walk Score emphasizes the absence of nearby amenities, while WalkCLIP highlights that some pedestrian facility exists at the site itself.
\section{Limitations \& Future Work}
WalkCLIP improves walkability prediction over baseline models, but limitations remain. First, the image–caption pairs for contrastive pretraining were generated with GPT-4o. These captions can contain inconsistent details, which introduces noise into the training signal and weakens image–text alignment. Future work should test human-curated captions to strengthen semantic correspondence between text and imagery. Second, the model relies on static satellite and street view imagery combined with aggregated non-visual features. Walkability, however, is dynamic and varies with time of day, season, and socio-demographic context. Incorporating temporal signals such as time-stamped mobility data, seasonal satellite imagery, or event-based changes could capture these variations effectively. Third, our evaluation is limited to a narrow set of cities. Expanding the analysis to urban areas with diverse geographic and socioeconomic characteristics would provide a robust test of model generalization across different contexts.
\section{Conclusion}
In this work, we introduced WalkCLIP, a multimodal framework for predicting urban walkability by combining satellite imagery, street view imagery, and population dynamics context. WalkCLIP aligns visual features with walkability-related semantics through contrastive learning and models spatial relationships with the SAFE module. Our results show that integrating top-down, ground-level, and behavioral perspectives yields more accurate walkability predictions than using any single modality alone. Future work should extend to a larger set of global cities, incorporate temporal signals, and refine training data with human-verified supervision.
\begin{acks}
We gratefully acknowledge the support of the Minnesota Undergraduate Research \& Academic Journal(MURAJ) In-Action Grant at the University of Minnesota.
\end{acks}
\balance
\bibliographystyle{ACM-Reference-Format}
\bibliography{references}
\end{document}